\documentclass[runningheads]{llncs}

\usepackage{eccv}

\usepackage{eccvabbrv}

\usepackage{graphicx}
\usepackage{booktabs}
\usepackage{multirow}
\usepackage{capt-of}
\usepackage{algorithm}
\usepackage{algpseudocode}

\usepackage{wrapfig}
\usepackage{caption}
\floatstyle{ruled}
\restylefloat{algorithm}

\algrenewcommand\algorithmicrequire{\textbf{Input:}}
\algrenewcommand\algorithmicensure{\textbf{Output:}}

\usepackage{tikz}
\usepackage[table]{xcolor}

\definecolor{goldmedal}{RGB}{255,215,0}
\definecolor{silvermedal}{RGB}{190,190,190}
\definecolor{bronzemedal}{RGB}{205,127,50}
\definecolor{rankgoldbg}{RGB}{255,249,210}
\definecolor{ranksilverbg}{RGB}{242,242,242}
\definecolor{rankbronzebg}{RGB}{250,238,228}

\newcommand{\goldrank}{%
  \tikz[baseline=(char.base)]{
    \node[
      circle,
      draw=black,
      fill=goldmedal,
      inner sep=1.1pt,
      line width=0.4pt
    ] (char) {\scriptsize\bfseries 1};
  }%
}

\newcommand{\silverrank}{%
  \tikz[baseline=(char.base)]{
    \node[
      circle,
      draw=black,
      fill=silvermedal,
      inner sep=1.1pt,
      line width=0.4pt
    ] (char) {\scriptsize\bfseries 2};
  }%
}

\newcommand{\bronzerank}{%
  \tikz[baseline=(char.base)]{
    \node[
      circle,
      draw=black,
      fill=bronzemedal,
      inner sep=1.1pt,
      line width=0.4pt
    ] (char) {\scriptsize\bfseries 3};
  }%
}

\usepackage{hyperref}

\usepackage{orcidlink}

\begin{document}

\title{KVeil: First-Place Solution to ECCV 2026 LifeGenIP Challenge on Unlearnable Videos}

\titlerunning{KVeil: First-Place Solution to ECCV 2026 LifeGenIP Challenge}

\author{{Boheng Li\inst{1}} \and
Yueming Xing\inst{2} \and
Yiming Li\inst{1}\thanks{Corresponding author. E-mail: \texttt{liyiming.tech@gmail.com}}\and
\linebreak
Jie Zhang\inst{3}\and
Han Qiu\inst{4}\and
Tianwei Zhang\inst{1}
}

\authorrunning{B.~Li et al.}

\institute{Nanyang Technological University, Singapore \and
Wuhan University, China \and
SiliconProspect AI \and
Tsinghua University, China
}

\maketitle

\begin{abstract}

The rapid development of personalized video generation has raised serious concerns over identity privacy, as a few facial videos of a person now suffice to reproduce their identity in synthetic videos. This report presents the \textbf{first-place} solutions to both phases of the ECCV 2026 LifeGenIP Challenge on Unlearnable Videos, which aims to protect facial videos from unauthorized customization under a limited and imperceptible perturbation budget. For Phase 1, we introduce Reference Capacity Collapse, a pair of complementary losses that degrade the information carried by the reference condition while keeping subsequent video representations dependent on it, together with a Motion-Guided EOT mechanism that improves robustness across unknown inference conditions. Phase 2 further considers tuning-based customization and unseen generative models. To address both threats within a single perturbation, we design a two-layer framework that protects the facial and background regions separately, constructing structure-aligned perturbations from a synthetic target identity in the former and strengthening the white-box disruption of Phase 1 in the latter. Our method achieves nearly perfect protection under white-box reference-based customization, and remains effective under other customization paradigms, various models, and temporal attacks. Both solutions are built upon our recent work LETHE. We hope our solutions can inspire future work on protecting facial videos against unauthorized customization. Our code is available at \url{https://github.com/AntigoneRandy/KVeil}.

\end{abstract}

\section{Introduction}

Recent advances in video generation models~\cite{wan2025, hunyuanvideo2025, hacohen2025ltx2, kuaishou2026kling3, yang2024cogvideox, seedance2026seedance, zheng2025open} have enabled realistic and temporally coherent videos to be synthesized from text prompts alone. Beyond generating generic content, there is a growing interest in personalizing these models to produce videos of a specific identity. This is typically achieved by either finetuning a pre-trained model on a small set of videos of the target person~\cite{hunyuanvideo2025,hacohen2025ltx2}, or, in more recent reference-based paradigms, by conditioning the generation on a single reference frame at inference time~\cite{wan2025,yang2024cogvideox,seedance2026seedance}. As the underlying models continue to improve, the resulting videos have become increasingly realistic and faithful to the identity being imitated. However, this capability also poses serious risks: a handful of clips shared online, or even a single portrait, now suffice to synthesize convincing videos of an individual without their consent, giving rise to deepfakes and other harmful content.

One practical solution is to leverage adversarial attacks to render the released data unexploitable. The key idea is to proactively add small perturbations to videos before releasing them, such that they remain visually intact to human eyes while causing downstream personalization to fail. However, existing methods primarily target text-to-image generation~\cite{liang2023adversarial, van2023anti, wang2024simac, liu2024metacloak, shan2023glaze, wu2025myopia}, typically by crafting perturbations that maximize the denoising loss or degrade the features encoded by the Variational AutoEncoder (VAE). These methods transfer poorly to modern video generators, which adopt strong transformer-based architectures together with more advanced 3D VAEs and are therefore much more resistant to such perturbations. Moreover, existing methods are mostly designed to protect a single static image. How to effectively and efficiently protect a video, where the identity is distributed across multiple frames, remains largely unexplored.

\begin{table*}[t]
\centering
\caption{Leaderboard results of Phase 1 in the LifeGenIP Challenge. Higher values indicate better performance for all metrics. Invis., Eff., and Rob. refer to invisibility, effectiveness, and robustness, respectively.}
\label{tab:phase1_leaderboard}
\resizebox{0.65\textwidth}{!}{
\begin{tabular}{c l c c c c c c}
\toprule
\textbf{Rank} & \textbf{Team} & \textbf{Invis. $\uparrow$} & \textbf{Eff. $\uparrow$} &
\textbf{Rob.$_1$ $\uparrow$} & \textbf{Rob.$_2$ $\uparrow$} &
\textbf{Rob.$_3$ $\uparrow$} & \textbf{Final $\uparrow$} \\
\midrule

\rowcolor{rankgoldbg}
\goldrank & \textbf{KVeil} &
\textbf{55.70} & \textbf{80.03} & \textbf{79.95} &
\textbf{79.16} & \textbf{82.85} & \textbf{77.72} \\

\rowcolor{ranksilverbg}
\silverrank & JLShen &
56.98 & 82.69 & 84.29 & 85.55 & 53.18 & 71.63 \\

\rowcolor{rankbronzebg}
\bronzerank & Rover &
68.02 & 74.74 & 77.68 & 76.54 & 61.27 & 71.46 \\

\bottomrule
\end{tabular}}
\end{table*}

\begin{table*}[t]
\centering
\caption{Leaderboard results of Phase 2 in the LifeGenIP Challenge. Higher values indicate better performance for all metrics. Invis., Eff., Rob., and Trans. refer to invisibility, effectiveness, robustness, and transferability, respectively.}
\label{tab:phase2_leaderboard}
\resizebox{\textwidth}{!}{
\begin{tabular}{clcccccccccccc}
\toprule
\multirow{2}{*}{\textbf{Rank}} & \multirow{2}{*}{\textbf{Team}} &
\multicolumn{1}{c}{\textbf{Origin}} &
\multicolumn{5}{c}{\textbf{Reference-based}} &
\multicolumn{5}{c}{\textbf{Tuning-based}} &
\multirow{2}{*}{\textbf{Final $\uparrow$}} \\
\cmidrule(lr){3-3}\cmidrule(lr){4-8}\cmidrule(lr){9-13}
& & \textbf{Invis. $\uparrow$}
& \textbf{Eff. $\uparrow$} & \textbf{Rob$_1$ $\uparrow$} & \textbf{Rob$_2$ $\uparrow$}
& \textbf{Trans$_1$ $\uparrow$} & \textbf{Trans$_2$ $\uparrow$}
& \textbf{Eff. $\uparrow$} & \textbf{Rob$_1$ $\uparrow$} & \textbf{Rob$_2$ $\uparrow$}
& \textbf{Trans$_1$ $\uparrow$} & \textbf{Trans$_2$ $\uparrow$} & \\
\midrule

\rowcolor{rankgoldbg}
\goldrank & \textbf{KVeil} &
\textbf{38.99} & \textbf{50.44} & \textbf{62.74} & \textbf{31.90} &
\textbf{27.09} & \textbf{29.57} & \textbf{46.86} & \textbf{43.94} &
\textbf{34.19} & \textbf{38.38} & \textbf{49.97} & \textbf{38.88} \\

\rowcolor{ranksilverbg}
\silverrank & Pawn &
49.92 & 45.88 & 54.44 & 30.11 & 21.75 & 27.20 &
44.80 & 47.21 & 43.83 & 31.45 & 51.79 & 37.77 \\

\rowcolor{rankbronzebg}
\bronzerank & Sprint &
50.96 & 66.59 & 50.03 & 28.81 & 23.46 & 26.27 &
48.14 & 40.87 & 46.56 & 29.57 & 40.17 & 37.59 \\

\bottomrule
\end{tabular}}
\end{table*}

\begin{figure}[t]
    \centering
    \includegraphics[width=0.8\linewidth]{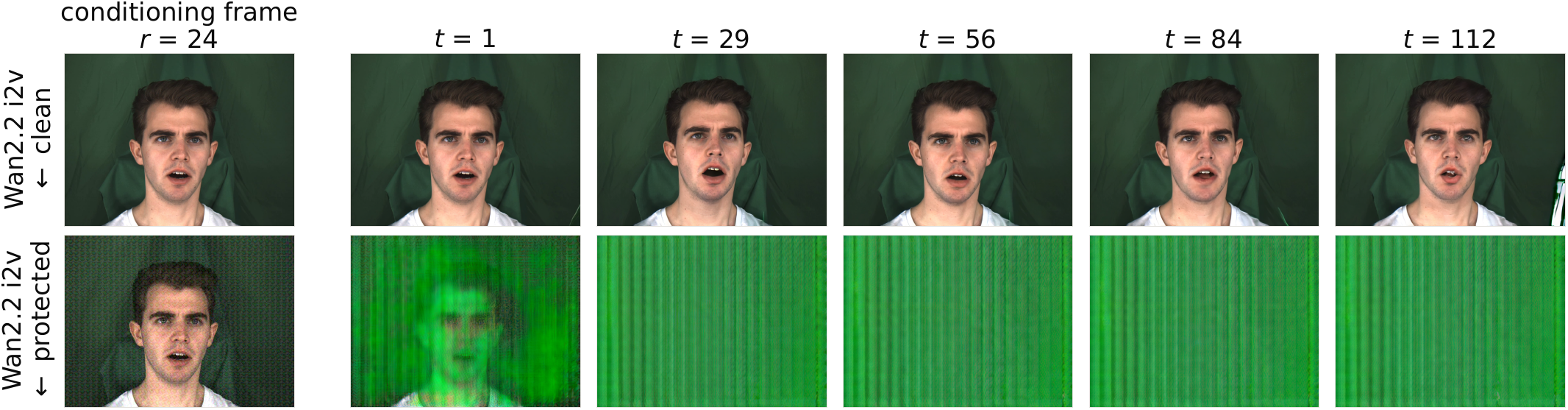}
    \caption{Qualitative results of Phase 1 under reference-based customization. }
    \label{fig:phase1_results}
\end{figure}

The ECCV 2026 LifeGenIP Challenge on Unlearnable Videos\cite{huang2026delving} establishes a systematic benchmark for evaluating video protection methods. Unlike prior evaluations that examine protection in isolated settings, this benchmark covers several practical dimensions. The challenge consists of two phases. Phase 1 focuses on reference-based customization, where the attacker may first apply undisclosed temporal processing to the protected clip, then sample a random frame and feed it together with an unknown text prompt to a known I2V model. Phase 2 considers both reference-based and tuning-based threats, where in the latter the attacker finetunes a video diffusion model on the full set of protected clips of one identity, and then prompts the customized model to generate new videos. Phase 2 additionally introduces more unseen temporal attacks and unseen base models to evaluate the transferability of the protection. We further elaborate the settings of the two phases in Section~\ref{sec:setting}.

Our team, KVeil, ranked first in both phases, as shown in Tables~\ref{tab:phase1_leaderboard} and~\ref{tab:phase2_leaderboard}. This report presents the details of our solutions for both phases. Both solutions are built upon LETHE~\cite{li2026lethe}, our recent study on how I2V models read the reference image and how this reading process can be disrupted for protection. This report documents how we adapt our methodology to the challenge settings, and we refer readers to our original paper for a more complete and in-depth scientific study, including its theoretical analysis and extensive evaluations. For Phase 1, we instantiate the two complementary losses of LETHE as a V-capacity collapse loss that disrupts the information-carrying capacity of the value tokens of the reference frame during self-attention, and a K-attract loss that keeps subsequent video representations dependent on this degraded condition, so that the model can hardly recover the target identity from the reference frame, as shown in Figure~\ref{fig:phase1_results}. Since the frame and prompts chosen by the attacker are unknown, we select a few representative reference frames by maximum-flow motion magnitude and optimize a shared perturbation over an expectation of frames, prompts, and input transformations. For Phase 2, we protect the facial and background regions separately: in the facial region we construct a structural perturbation from a synthetic target identity, so that the identity-level disruption persists through personalization and transfers well across models; in the background region we retain and strengthen the Phase 1 white-box disruption to maintain strong protection against the known model. The two components are optimized sequentially in a face-first, background-second manner, combining transferable identity-level protection with strong white-box disruption.

\section{Related Work}
\label{sec:related}

\subsection{Diffusion-based Customization in Image and Video Generation}

\textbf{Image Generation and Customization.}
Early text-to-image diffusion models, represented by Stable Diffusion~\cite{Rombach_2022_CVPR}, compress images into a lower-dimensional latent space and generate within it using a UNet-based denoising network. Built upon such models, subject-driven customization develops a method to teach a pre-trained model to generate a specific concept. These methods are commonly training-based~\cite{hu2021lora}: by updating the entire model or a small subset of its parameters, they bind the target subject to a special token, after which the subject can be reproduced by including the token in the prompt. Subsequent studies have developed more efficient and fine-grained customization methods, achieving better subject fidelity with fewer reference images and lower training cost~\cite{gal2022image, ruiz2023dreambooth, kumari2023multi}.

\noindent\textbf{Video Generation and Customization.}
As generative capabilities extended from images to videos, the underlying architecture has also changed considerably. Modern video generation models~\cite{wan2025, yang2024cogvideox, hunyuanvideo2025, hacohen2025ltx2, kuaishou2026kling3} first compress a video along both spatial and temporal dimensions using a 3D VAE pre-trained on large-scale data, and then generate within this latent space using a Diffusion Transformer (DiT). Compared with the earlier UNet, DiT relies on a considerably larger number of attention layers to integrate and extract information: the frame features and text conditions are repeatedly read and aggregated across these layers and gradually propagated into the whole spatiotemporal representation, with the attention mechanism selecting the information that is useful for the current generation.

The way customization is performed has extended accordingly. Besides traditional tuning-based customization, an increasing number of video models directly support reference-based generation~\cite{wan2025, seedance2026seedance}, where a single reference frame is fed to the model as a condition to generate a new video that preserves the same identity. As the underlying models improve, such generations have become increasingly realistic, while requiring neither training nor more than one image from the target subject. Since it requires no training and only a single frame, reference-based generation has also become a standard service offered by many video generation models~\cite{wan2025,yang2024cogvideox} and online services~\cite{kuaishou2026kling3,seedance2026seedance}.

\subsection{Protection against Unauthorized Customization}

\textbf{Protection against Image Customization.}
As diffusion-based personalization became widely available, the privacy risks it brings have drawn increasing attention, and a line of work has started to develop proactive defenses. A popular approach is to add adversarial perturbations to images to render them \textit{unlearnable} before release, such that tuning-based customization either learns incorrect features or fails to converge. This is typically achieved along two routes. One is to disrupt the training objective of the diffusion model, maximizing the denoising loss on the protected images so that the training is derailed or misled~\cite{liang2023adversarial, van2023anti}. The other is to disrupt the latent representation of the image, pushing the VAE-encoded features of the protected image away from those of the original one, so that the visual information the model receives is already degraded~\cite{shan2023glaze, wu2025myopia}. Later studies improve the robustness of such perturbations, optimizing over diffusion timesteps, surrogate training dynamics, and different prompts, so as to generalize better across models and preprocessing conditions~\cite{wang2024simac, liu2024metacloak}.

\noindent\textbf{Protection against  Video Customization.}
With the adoption of I2V generation, similar ideas have been extended to video models. Existing methods typically target reference-based generation and inherit the routes above~\cite{pang2024vgmshield, gui2025i2vguard, chowdhury2025vid, vu2026anti, guo2026i2vshield}. These methods, however, are mostly effective on earlier video models extending from the UNet era, and their effect on latest DiT-based models is rather limited.

On one hand, the 3D VAE adopted by modern video models is pre-trained on large-scale data and offers stronger compression and reconstruction, so that adversarial noise under a limited budget can hardly alter the latent features to a meaningful extent. On the other hand, methods that aim to disrupt the training process are even less effective, since reference-based generation injects the reference frame merely as a condition at inference time and involves no parameter update at all. The abundant attention layers in DiT further exacerbate the situation. As described earlier, the conditioning information is repeatedly read and aggregated across a large number of attention layers, and the attention mechanism keeps selecting the information that is useful for the current generation. Thus, adversarial perturbations are largely treated as noise and discarded during this propagation, leaving little impact in the final video. Besides, existing I2V protection mostly takes a single reference image as the object to protect, rather than a complete portrait video. Perturbations optimized naively over an entire video tend to be fragile and can be removed by simple temporal attacks such as frame averaging. Therefore, how to protect videos against video customization remains an open challenge and is worthy of further exploration.

\section{Proposed Approach}
\label{sec:setting}

\textbf{Challenge Overview.} The LifeGenIP Competition requires participants to add imperceptible perturbations to portrait videos to prevent unauthorized diffusion-based video customization. We summarize the evaluation settings in Table~\ref{tab:challenge_overview}.

\begin{table*}[t]
\centering
\caption{Evaluation protocols in the two phases. Each check mark refers to one evaluated setting. FIP combines face, image quality, and prompt following scores. Invis. refers to invisibility score. Gray text indicates black-box evaluation settings.}
\label{tab:challenge_overview}
\small
\setlength{\tabcolsep}{2pt}
\renewcommand{\arraystretch}{1.1}
\resizebox{0.85\textwidth}{!}{
\begin{tabular}{p{1.55cm} p{2.2cm} p{1.4cm} p{4.6cm} p{4.4cm}}
\toprule
\textbf{Attack Pathways} & \textbf{Track} & \textbf{Metrics} & \textbf{Phase 1} & \textbf{Phase 2} \\
\midrule
& Invisibility & Invis. & $\checkmark$ & $\checkmark$ \\
\midrule
\multirow{3}{*}[-1.8em]{\shortstack[l]{Reference-\\based}}
& Effectiveness & FIP & $\checkmark$~Wan2.2-TI2V-5B & $\checkmark$~Wan2.2-TI2V-5B \\
\cmidrule(l){2-5}
& Robustness & FIP
& \textcolor{gray}{$\checkmark$~Neighbor-frame interpolation}\newline
  \textcolor{gray}{$\checkmark$~7-frame temporal averaging}\newline
  \textcolor{gray}{$\checkmark$~H.264 re-compression}
& \textcolor{gray}{$\checkmark$~7-frame temporal averaging + H.264 re-compression}\newline
  \textcolor{gray}{$\checkmark$~VideoPure purification} \\
\cmidrule(l){2-5}
& Transferability & FIP
& \textcolor{gray}{--}
& \textcolor{gray}{$\checkmark$~HunyuanVideo-1.5}\newline
  \textcolor{gray}{$\checkmark$~Kling 3.0} \\
\midrule
\multirow{3}{*}[-1.8em]{\shortstack[l]{Tuning-\\based}}
& Effectiveness & FIP
& --
& $\checkmark$~Wan2.2-TI2V-5B \\
\cmidrule(l){2-5}
& Robustness & FIP
& \textcolor{gray}{--}
& \textcolor{gray}{$\checkmark$~5-frame temporal averaging + temporal reversal}\newline
  \textcolor{gray}{$\checkmark$~H.264 re-compression} \\
\cmidrule(l){2-5}
& Transferability & FIP
& \textcolor{gray}{--}
& \textcolor{gray}{$\checkmark$~LTX-2}\newline
  \textcolor{gray}{$\checkmark$~HunyuanVideo-1.5} \\
\bottomrule
\end{tabular}
}
\end{table*}

\noindent\textbf{Attack pathways.}
The competition considers two customization settings:
\begin{itemize}
    \item Reference-based customization. The attacker samples one frame from a protected clip and uses it with a text prompt as input to an I2V model.
    \item Tuning-based customization. The attacker uses the protected clips of one identity to finetune a video diffusion model with LoRA, and then generates new videos from text prompts.
\end{itemize}

\noindent\textbf{Evaluation tracks.}
Protection is evaluated from three aspects:
\begin{itemize}
    \item Effectiveness. Protected videos are used for generation with the known target model, Wan2.2-TI2V-5B\cite{wan2025}, and the generated videos are then evaluated.
    \item Robustness. Protected videos first undergo hidden temporal processing or purification, after which they are used for generation and evaluated.
    \item Transferability. Protected videos are used for generation with unseen black-box video generation models, and the generated videos are then evaluated.
\end{itemize}

\noindent\textbf{Evaluation metrics.}
The competition reports four scores:
\begin{itemize}
    \item Invisibility measures similarity between the original and protected videos, with higher values indicating less perceptible perturbations.
    \item Face Score measures the disruption of the identity in
generated videos, where stronger protection corresponds to lower face detection or identity similarity.
    \item Prompt Following Score measures consistency with the attack prompt, with stronger disruption corresponding to lower prompt adherence.
    \item Image Quality Score measures the visual quality of generated videos, with stronger disruption corresponding to lower perceptual quality.
\end{itemize}

The official evaluator converts these measurements into protection scores, which range from 0 to 100, with higher values consistently indicating stronger protection. In the following sections, we describe our methods for both phases.

\subsection{Phase 1}

Phase 1 focuses on reference-based video customization. Given a 113-frame, $640\times480$ facial video\cite{kaisiyuan2020mead}, we add perturbations under an $\ell_\infty\leq16/255$ constraint to prevent sampled reference frames from enabling personalized video generation.

Accordingly, Phase 1 contains two main components. \textbf{Reference Capacity Collapse} directly disrupts how the reference condition is represented and propagated within Wan2.2-TI2V-5B. \textbf{Motion-Guided EOT Optimization} jointly considers reference sampling uncertainty and other unknown inference conditions to improve robustness under the hidden evaluation settings.

\paragraph{\textbf{Reference Capacity Collapse.}}
This component directly inherits the core mechanism of LETHE~\cite{li2026lethe}, which identifies the diversity of reference Values as a key factor governing how I2V models read the reference image.

In Wan2.2-TI2V-5B\cite{wan2025}, the VAE-encoded reference image occupies the first temporal position of the video latent, while the remaining positions are initialized with noise. During denoising, these latents are patchified into a shared spatiotemporal token sequence, where reference and future video tokens interact through global self-attention\cite{guo2026i2vshield, chowdhury2025vid}. Thus, perturbing the initial VAE latent alone does not directly limit the information that reference representations provide during subsequent spatiotemporal modeling.

We address this issue from two complementary aspects of self-attention. \textbf{V-capacity collapse} reduces the spatial diversity of reference-conditioned Values, making their fine-grained identity representations more homogeneous and reducing the effective capacity of the reference region. \textbf{K-attract} then increases the matching preference of future temporal Queries toward this degraded reference region, encouraging subsequent representations to remain influenced by the corrupted condition. Together, they directly reduce the information available from the reference while still maintaining its influence on later generation.

\noindent\underline{\textbf{V-capacity collapse.}}
Values determine the visual information that the reference frame provides to subsequent frames. Different spatial tokens normally encode distinctive local appearance features. We reduce their variation to further compress the representation capacity of the reference frame.
Let $V_0^{h}=\{v_{0,i}^{h}\}_{i=1}^{N_0}$ denote the Value tokens of the $h$-th attention head in the reference temporal region. We compute the normalized dispersion within each head and average it over all $H$ heads, so that heads with different value scales contribute equally:
\begin{equation}
\bar v_0^{h}
=
\frac{1}{N_0}\sum_{i=1}^{N_0}v_{0,i}^{h},
\qquad
\mathcal{L}_{V}
=
\frac{1}{H}\sum_{h=1}^{H}
\frac{
\sum_{i=1}^{N_0}\|v_{0,i}^{h}-\bar v_0^{h}\|_2^2
}{
\sum_{i=1}^{N_0}\|v_{0,i}^{h}\|_2^2+\xi
}.
\end{equation}
Minimizing $\mathcal{L}_{V}$ reduces the spatial dispersion of reference Values, making different tokens more homogeneous for fine-grained identity features.

\noindent\underline{\textbf{K-attract.}}
Capacity collapse alone does not ensure that the degraded reference continues to influence later generation, as subsequent latents gradually form their own representations. Thus, we use the Keys to encourage future frame Queries to keep retrieving from the reference region.

For the $h$-th attention head, we average-pool the future-frame Queries and reference-region Keys over token indices $\mathcal{I}_{>0}$ and $\mathcal{I}_{0}$, respectively, and define
\begin{equation}
\begin{aligned}
\bar q_{>0}^{\,h}
&=
\frac{1}{N_{>0}}
\sum_{i\in\mathcal{I}_{>0}}q_i^h,
\qquad
\bar k_{0}^{\,h}
=
\frac{1}{N_0}
\sum_{j\in\mathcal{I}_{0}}k_j^h,\\
\mathcal{L}_{K}
&=
-\frac{1}{H}
\sum_{h=1}^{H}
\frac{
\left\langle
\operatorname{sg}(\bar q_{>0}^{\,h}),
\bar k_{0}^{\,h}
\right\rangle
}{
\sqrt{d_h}
},
\end{aligned}
\end{equation}
where $H$ denotes the number of attention heads, $d_h$ denotes the dimension of each head, and $\operatorname{sg}(\cdot)$ denotes stop-gradient.
Minimizing $\mathcal{L}_{K}$ pulls the reference Keys toward the current retrieval directions of future Queries, encouraging later frames to keep retrieving from the reference.

\noindent\underline{\textbf{Auxiliary Objectives.}}
V-capacity collapse and K-attract constitute the main protection mechanism by directly disrupting identity information in the reference representation and its propagation. To further improve performance across all evaluation metrics, we introduce several complementary objectives to strengthen the resulting perturbation. These objectives additionally perturb the VAE representation, reduce text-conditioned feature diversity, and suppress high-quality facial reconstruction. Their detailed formulations are provided in Appendix~\ref{aux}.

\paragraph{\textbf{Motion-Guided EOT Optimization.}}
The objectives above specify how the perturbation disrupts the model, while the reference frame and inference conditions used by the attacker remain unknown. Optimizing under a fixed reference frame and inference configuration may therefore lead to overfitting. We address this uncertainty with Expectation-over-Transformation (EOT), jointly randomizing reference frames, input transformations, prompts, and Transformer depths.

\noindent\underline{\textbf{Motion-Guided Video-Level Shared Perturbation.}}
Optimizing full video is computationally expensive, while independently protecting each frame produces temporally inconsistent perturbations that are vulnerable to temporal processing. We therefore learn a single video-level perturbation $\delta$ from a small set of representative frames and apply it to all frames:
\begin{equation}
\tilde{x}_t
=
\operatorname{Clip}(x_t+\delta),
\qquad
t=1,\ldots,T,
\qquad
\|\delta\|_{\infty}\leq\epsilon.
\end{equation}
The shared perturbation preserves temporal consistency, while representative frames encourage generalization across different subject states.
To select representative frames, we estimate dense optical flow between adjacent frames using Farneb\"ack optical flow\cite{farneback2003two} and define the motion magnitude as
\begin{equation}
F_t=\operatorname{Flow}(x_{t-1},x_t),\qquad m_t=\frac{1}{|\Omega|}\sum_{p\in\Omega}\|F_t(p)\|_2 ,
\end{equation}
where $m_t$ measures the motion of frame $t$ relative to the preceding frame. Larger values correspond to stronger pose, expression, or facial-structure changes, making these frames more informative for learning the shared perturbation.

Using only the maximum-motion frame covers a single instantaneous state, while selecting the global Top-$K$ frames may concentrate samples around the same motion segment. We therefore partition the video into $K$ motion-energy-balanced windows according to normalized cumulative motion and select the maximum-motion frame from each window:
\begin{equation}
t_k^{*}=\arg\max_{t\in\mathcal{W}_k}m_t,\quad k=1,\ldots,K,\qquad \mathcal{S}=\{x_{t_k^{*}}\}_{k=1}^{K}.
\end{equation}

This strategy assigns comparable motion energy to each window. We set $K=5$ and randomly sample reference frames from $\mathcal S$ to jointly optimize the same shared perturbation $\delta$, improving generalization to unseen reference frames.

\noindent\underline{\textbf{EOT Optimization.}}
EOT\cite{athalye2018synthesizing} has been widely adopted to improve the robustness of adversarial perturbations against input transformations\cite{liu2024metacloak,zhou2025dormant}. Following this, we randomize input preprocessing, text prompts, and Transformer depth to reduce dependence on a fixed inference configuration. We denote an EOT instance as $\omega=(x,\tau,p,b)$, where $x\sim\mathcal S$, $\tau\sim\mathcal T$, $p\sim\mathcal P$, and $b\sim\mathcal B$. $\mathcal S$ contains five representative frames, $\mathcal T$ includes identity, JPEG compression, Gaussian blur, Gaussian noise, and random downsample-upsample, $\mathcal P$ contains six semantically similar prompts, and $\mathcal B=\{0,3,6\}$ contains the target Transformer blocks. At each optimization step $s$, we estimate the EOT objective with $M$ samples:

\begin{equation}
\widehat{\mathcal L}_{s}(\delta)=\frac{1}{M}\sum_{r=1}^{M}\mathcal L_{\mathrm{total}}(\delta;\omega_r),\qquad \omega_r\sim\mathcal S\times\mathcal T\times\mathcal P\times\mathcal B.
\end{equation}
This exposes the shared perturbation to different inference conditions, improving robustness and cross-frame generalization under unknown evaluation settings.

\subsubsection{Overall Optimization.}
We combine all objectives and optimize the shared perturbation with EOT:
\begin{equation}
\mathcal L_{\mathrm{total}}
=
\mathcal L_V+\lambda_K\mathcal L_K+\sum\nolimits_{j}\lambda_j\mathcal L_j,
\quad
\delta^{*}
=
\arg\min_{\|\delta\|_{\infty}\leq\epsilon}
\mathbb E_{\omega\sim
\mathcal S\times\mathcal T\times\mathcal P\times\mathcal B}
\left[
\mathcal L_{\mathrm{total}}(\delta;\omega)
\right].
\label{eq:overall_objective}
\end{equation}
where $\mathcal L_j$ denotes the $j$-th auxiliary objective with weight $\lambda_j$ (see Appendix~\ref{aux}). We summarize the overall optimization procedure in Algorithm~\ref{alg:phase1_optimization}.

\begin{figure}[t]
\centering

\begin{minipage}[t]{\linewidth}\small
\hrule
\refstepcounter{algorithm}
\noindent
\rule[-0.8ex]{0pt}{3.4ex}%
\textbf{Algorithm~\thealgorithm}\enspace Phase 1 Pipeline
\label{alg:phase1_optimization}
\hrule

\begin{algorithmic}[1]
\Require white-box I2V model $\mathcal F_\theta$, $\mathcal S$, $\mathcal T$, $\mathcal P$, $\mathcal B$, optimization steps $S$, EOT samples $M$, perturbation budget $\epsilon$,
\Ensure Universal perturbation $\delta$

\State Pre-compute text embeddings for all $p\in\mathcal P$

\State $\delta
\sim
\mathcal U(-\epsilon/4,\epsilon/4)$,
$m\gets0$

\For{$s=1,\ldots,S$}

    \State Clear accumulated gradient of $\delta$

    \For{$r=1,\ldots,M$}

        \State Sample
        $x_j\sim\mathcal S$,
        $\tau\sim\mathcal T$,
        $p\sim\mathcal P$,
        $b\sim\mathcal B$

        \State $\tilde{x}_j
        \gets
        \tau\!\left(
        \operatorname{Clip}(x_j+\delta)
        \right)$

        \State $\tilde{z}_j
        \gets
        \operatorname{VAEEnc}(\tilde{x}_j)$

        \State Run I2V denoising to the target timestep
        using $\tilde{z}_j$ as reference condition

        \State Forward through Transformer blocks up to $b$

        \State Capture self-attention $(Q,K,V)$

        \State Compute
        $\mathcal L_V$,
        $\mathcal L_K$, and auxiliary objectives

        \State Accumulate
        $\nabla_{\delta}\mathcal L_{\mathrm{total}}/M$

    \EndFor

    \State $\delta \gets
    \operatorname{Update}
    \left(
    \delta,
    \nabla_\delta\mathcal L_{\mathrm{total}}
    \right)$

    \State $\delta \gets
    \Pi_{\|\delta\|_\infty\leq\epsilon}(\delta)$

\EndFor

\State \Return $\delta$

\end{algorithmic}
\hrule
\end{minipage}

\end{figure}

\subsection{Phase 1 Experiments}
We evaluate our protection on Wan2.2-TI2V-5B using frame $r=24$, which is neither the first frame $f_0$ nor any frame in the representative set $\mathcal{S}$.

As shown in Figure~\ref{fig:phase1_results}, the original conditioning frame produces a temporally coherent video with preserved appearance and facial structure. With the protected frame, the subject becomes distorted and develops spatial and temporal artifacts, despite the conditioning frame itself remaining visually recognizable.

\subsection{Phase 2}
Phase 2 extends protection from a single video to a set of videos of the same identity. Participants protect 10 face-centric clips, each containing 113 frames at $640\times480$, under the same $\ell_\infty\leq16/255$ budget. Phase 2 considers both reference-based and tuning-based customization.
Both settings are evaluated in terms of Effectiveness, Robustness, and Transferability, together with the invisibility of the protected videos. We retain white-box access to Wan2.2-TI2V-5B, which is used for Effectiveness evaluation and can therefore guide perturbation optimization. The final prompts, reference sampling strategies, evaluators, temporal attacks, and transfer models remain hidden.

Compared with Phase 1, Phase 2 requires protection to generalize across paradigms and unseen models. We address this with a two-layer framework consisting of \textbf{Prior-Aligned Face Protection} and \textbf{White-Box Background Protection}, optimized sequentially and combined into the video perturbation.

\subsubsection{Prior-Aligned Face Protection.}

We observe that Phase 1 perturbations mainly contain high-frequency patterns with limited correlation to facial structure. Such signals can be weakened during black-box generation and are less likely to be consistently learned as identity-related features during tuning-based customization. We therefore introduce facial structure into the perturbation design.
Specifically, we use a synthetic target face $x_{\mathrm{tar}}$ with an identity different
\begin{figure}[t]
\centering
\begin{minipage}[t]{0.54\linewidth}\footnotesize
\hrule
\refstepcounter{algorithm}\label{alg:face_protection}
\noindent\raggedright\rule{0pt}{2.6ex}\textbf{Algorithm~\thealgorithm}\ Prior-Aligned Face Protection\rule[-0.9ex]{0pt}{0.9ex}\par
\hrule
\begin{algorithmic}[1]
\Require
$\mathcal X=\{x_t\}_{t=1}^{T}$,
$x_{\mathrm{tar}}$, $a$, $\epsilon$

\State
$m^{\mathrm{soft}},m^{\mathrm{hard}}
\gets
\operatorname{FaceMasks}(\mathcal X)$

\State
$\mathcal P_a
\gets
\operatorname{Landmark68}(x_a)$

\State
$\mathcal P_{\mathrm{tar}}
\gets
\operatorname{Landmark68}(x_{\mathrm{tar}})$

\State
$x_{\mathrm{tar}}^{a}
\gets
\operatorname{AlignToAnchor}
(x_{\mathrm{tar}},
\mathcal P_{\mathrm{tar}},
\mathcal P_a)$

\State
$d
\gets
\mathbf w\odot
\left[
\Phi_{\mathrm{Lab}}(x_{\mathrm{tar}}^{a})
-
\Phi_{\mathrm{Lab}}(x_a)
\right]$

\State
$d_{\mathrm{str}}
\gets
d-G_\sigma(d)$

\State
$r
\gets
\Phi_{\mathrm{Lab}}^{-1}
\left(
\Phi_{\mathrm{Lab}}(x_a)
+\gamma d_{\mathrm{str}}
\right)-x_a$

\State
$\delta_{\mathrm{face}}^{a}
\gets
\epsilon m^{\mathrm{soft}}
\odot
\operatorname{sign}(r)$

\State
$\{\mathcal P_t\}_{t=1}^{T}
\gets
\operatorname{Landmark68}(\mathcal X)$

\For{$t=1,\ldots,T$}
    \State
    $\delta_{\mathrm{face}}^t
    \gets
    \operatorname{TPSWarp}
    (\delta_{\mathrm{face}}^a;
    \mathcal P_a,\mathcal P_t)$

    \State
    $m_t
    \gets
    \operatorname{TPSWarp}
    (m^{\mathrm{hard}};
    \mathcal P_a,\mathcal P_t)$
\EndFor

\State
\Return
$\{\delta_{\mathrm{face}}^{t},m_t\}_{t=1}^{T}$

\end{algorithmic}
\hrule
\end{minipage}\hfill
\begin{minipage}[t]{0.445\linewidth}\footnotesize
\hrule
\refstepcounter{algorithm}\label{alg:layered_video_protection}
\noindent\raggedright\rule{0pt}{2.6ex}\textbf{Algorithm~\thealgorithm}\ Phase 2 Pipeline\rule[-0.9ex]{0pt}{0.9ex}\par
\hrule
\begin{algorithmic}[1]
\Require
$\mathcal X=\{x_t\}_{t=1}^{T}$,
$x_{\mathrm{tar}}$, $a$, $\mathcal F_\theta$, $\epsilon$

\State $\{\delta_{\mathrm{face}}^{t},m_t\}_{t=1}^{T}\gets$
\Statex \hspace{\algorithmicindent}$\operatorname{FaceProtect}(\mathcal X,x_{\mathrm{tar}},a,\epsilon)$

\State
$\mathcal S
\gets
\operatorname{MotionSample}
\left(
\mathcal X
\right)$

\State
$\Delta_{\mathrm{face}}
\gets
\{
m_t\odot\delta_{\mathrm{face}}^{t}
\}_{t=1}^{T}$

\State
$\mathcal M_{\mathrm{bg}}
\gets
\{1-m_t\}_{t=1}^{T}$

\State $\delta_{\mathrm{bg}}^{*}\gets\operatorname{BackgroundProtect}\big(\mathcal S,\mathcal F_\theta,$
\Statex \hspace{\algorithmicindent}$\Delta_{\mathrm{face}},\mathcal M_{\mathrm{bg}},\epsilon\big)$

\For{$t=1,\ldots,T$}
    \State $\delta_t\!\gets\! m_t{\odot}\delta_{\mathrm{face}}^{t}{+}(1{-}m_t){\odot}\delta_{\mathrm{bg}}^{*}$

    \State
    $\delta_t
    \gets
    \operatorname{Clip}_{[-\epsilon,\epsilon]}
    (\delta_t)$

    \State
    $\tilde{x}_t
    \gets
    \operatorname{Clip}(x_t+\delta_t)$
\EndFor

\State
$\widetilde{\mathcal X}
\gets
\{\tilde{x}_t\}_{t=1}^{T}$

\State
\Return $\widetilde{\mathcal X}$

\end{algorithmic}
\hrule
\end{minipage}
\end{figure}
from that of the protected subject, and use their appearance difference to construct a structured perturbation.

We generate $x_{\mathrm{tar}}$ with ChatGPT Image\cite{openai_chatgpt} while maintaining a similar frontal pose and imaging conditions, and select a result with a distinct identity and stable facial landmarks. We further evaluate the sensitivity of our method to different target
identities in Section~\ref{sensitive-target}.

As the target and protected faces differ in geometry, we detect the face \cite{guo2021sample} and extract facial landmarks\cite{insightface}. We apply Umeyama similarity transformation\cite{umeyama1991least} for global alignment and TPS warping\cite{bookstein1989principal} for local deformation:
\begin{equation}
x_{\mathrm{tar}}^{a}
=
\mathcal{A}(x_{\mathrm{tar}};\mathcal{P}_{\mathrm{tar}},\mathcal{P}_{a}),
\qquad
\mathcal{A}
=
\mathcal{W}_{\mathrm{TPS}}\circ\mathcal{W}_{\mathrm{sim}},
\end{equation}
where $\mathcal{P}_{\mathrm{tar}}$ and $\mathcal{P}_{a}$ are the landmarks of the target face and the anchor frame, respectively.
We then compute their appearance difference in CIELAB space and remove smooth illumination and color components:
\begin{equation}
d
=
\mathbf{w}\odot
\left[
\Phi_{\mathrm{Lab}}(x_{\mathrm{tar}}^{a})
-
\Phi_{\mathrm{Lab}}(x_a)
\right],
\qquad
d_{\mathrm{str}}
=
d-G_{\sigma}(d),
\end{equation}
where $\Phi_{\mathrm{Lab}}(\cdot)$ denotes the RGB-to-CIELAB transformation,
$G_{\sigma}(\cdot)$ denotes Gaussian smoothing with standard deviation
$\sigma$, $\mathbf{w}=(w_L,w_a,w_b)$ weights the CIELAB channels, and
$\gamma$ is a residual scaling coefficient. We then map the residual to RGB space and restrict it to a feathered face-region mask $m^{\mathrm{soft}}$:
\begin{equation}
\delta_{\mathrm{face}}^{a}
=
\epsilon m^{\mathrm{soft}}\odot
\operatorname{sign}
\left[
\Phi_{\mathrm{Lab}}^{-1}
\left(
\Phi_{\mathrm{Lab}}(x_a)+\gamma d_{\mathrm{str}}
\right)
-x_a
\right].
\end{equation}
The target identity serves as a structural reference for perturbation directions, aligning them with facial geometry and local texture.

We construct video-level face masks from all frames to cover the facial motion
\begin{wrapfigure}[12]{r}{0.52\textwidth}
    \centering
    \includegraphics[
        width=\linewidth
    ]{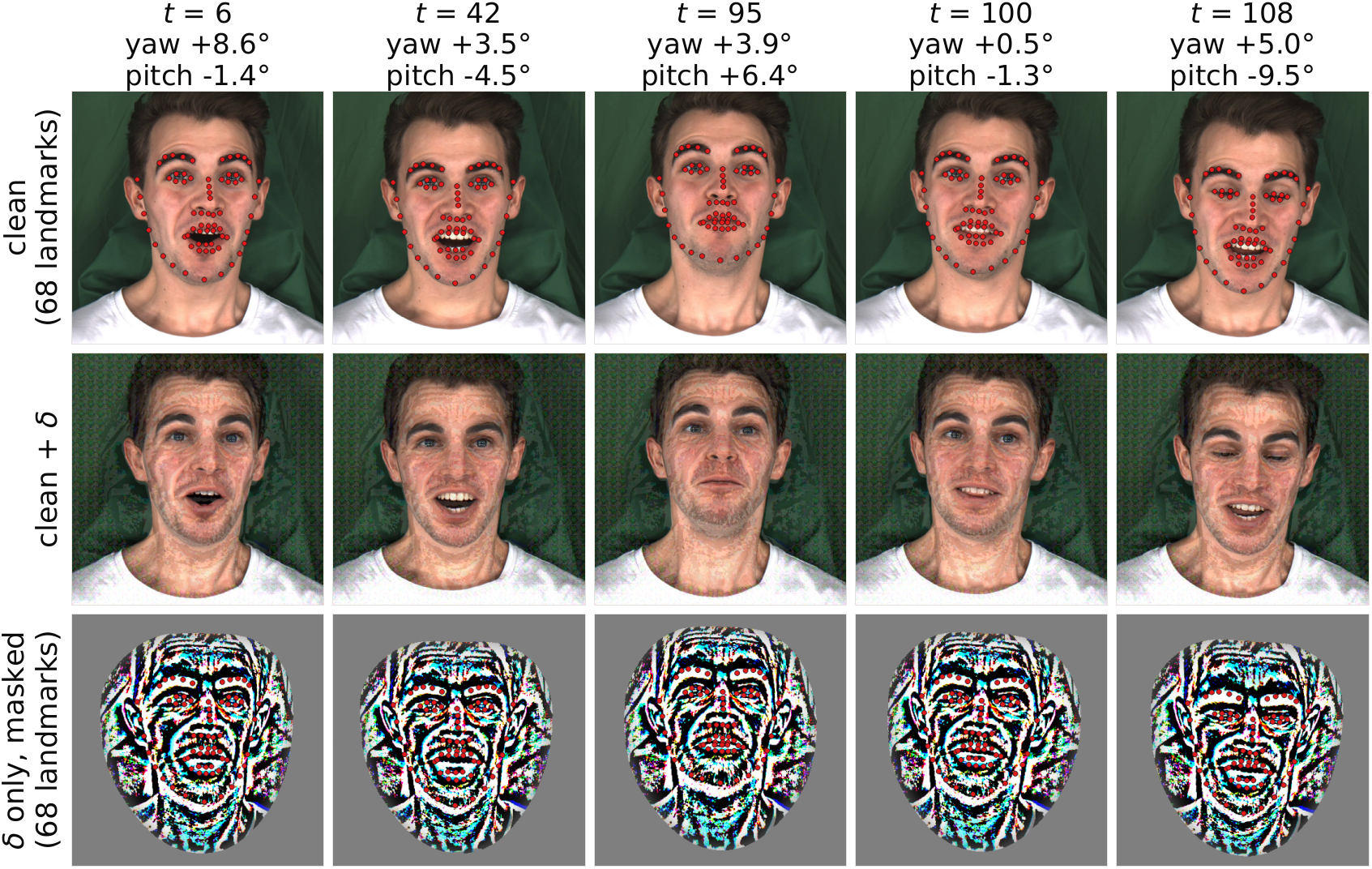}
    \caption{
        Landmark-guided propagation of the face-layer perturbation
        across different poses.
    }
    \label{fig:face_layer}
\end{wrapfigure}
range throughout the clip, obtaining a feathered $m^{\mathrm{soft}}$ and its binary counterpart $m^{\mathrm{hard}}$. To propagate the anchor perturbation over time, we estimate a landmark-guided TPS deformation between anchor frame and each frame and warp the perturbation and face-region mask to the corresponding facial geometry:
\begin{equation}
\begin{aligned}
\delta_{\mathrm{face}}^{t}
&=
\mathcal{W}
(\delta_{\mathrm{face}}^{a};
\mathcal P_a,\mathcal P_t),\\
m_t
&=
\mathcal{W}
(m^{\mathrm{hard}};
\mathcal P_a,\mathcal P_t).
\end{aligned}
\end{equation}
This allows the perturbation to follow facial motion across frames.

Algorithm~\ref{alg:face_protection} summarizes the Prior-Aligned Face Protection procedure. Figure~\ref{fig:face_layer} visualizes its propagation across different poses and expressions. The detected 68-point landmarks guide the alignment, and the perturbation remains associated with the corresponding facial regions over time.

\subsubsection{White-Box Background Protection.}

After constructing the prior-aligned
perturbation over the facial region, we exploit the remaining non-face region to preserve strong white-box effectiveness. We adapt the Phase 1 framework to the non-face region as a complementary background strategy.

However, reserving the facial region reduces the spatial area available for white-box optimization, especially around the identity-critical central region. To maintain effective disruption under this restricted optimization space, we retain the three objectives that directly affect reference encoding and propagation: V-capacity collapse, K-attract, and VAE latent disruption, while removing the remaining auxiliary objectives used in Phase 1. The resulting background perturbation complements the transferable face protection with strong reference-based disruption against the known model.

\subsubsection{Spatiotemporal Layer Coupling.}

We couple the two protection layers sequentially using a face-first, background-second strategy.
We first construct the prior-aligned face perturbation on the anchor frame and propagate it across frames using facial landmarks, ensuring consistent protected conditioning. We then retain the fixed face perturbation during white-box forward passes and optimize a shared background perturbation only over the non-face region. We combine the two layers using the frame-wise face masks under the shared budget $\ell_\infty\leq16/255$.
Algorithm~\ref{alg:layered_video_protection} summarizes the overall Phase 2 pipeline.

\subsection{Phase 2 Experiments}

\begin{table}[t]
\centering
\caption{Quantitative results of Phase 2.}
\label{tab:phase2_qualitative_eval}
\small
\begin{tabular}{lccc}
            \toprule
            \textbf{Customization Setting}
            & \textbf{Face $\uparrow$}
            & \textbf{IQ $\uparrow$}
            & \textbf{PF $\uparrow$} \\
            \midrule
            Wan2.2 I2V (Clean Ref.)
            & 16.79 & 25.39 & 12.33 \\
            Wan2.2 I2V (Protected Ref.)
            & 82.66 & 27.59 & 35.51 \\
            HunyuanVideo-1.5 I2V (Protected Ref.)
            & 34.40 & 26.08 & 26.53 \\
            Wan2.2 LoRA-1500 (Tune)
            & 51.09 & 37.11 & 17.64 \\
            \bottomrule
        \end{tabular}
\end{table}

\begin{figure}[t]
\centering
\includegraphics[width=0.6\linewidth]{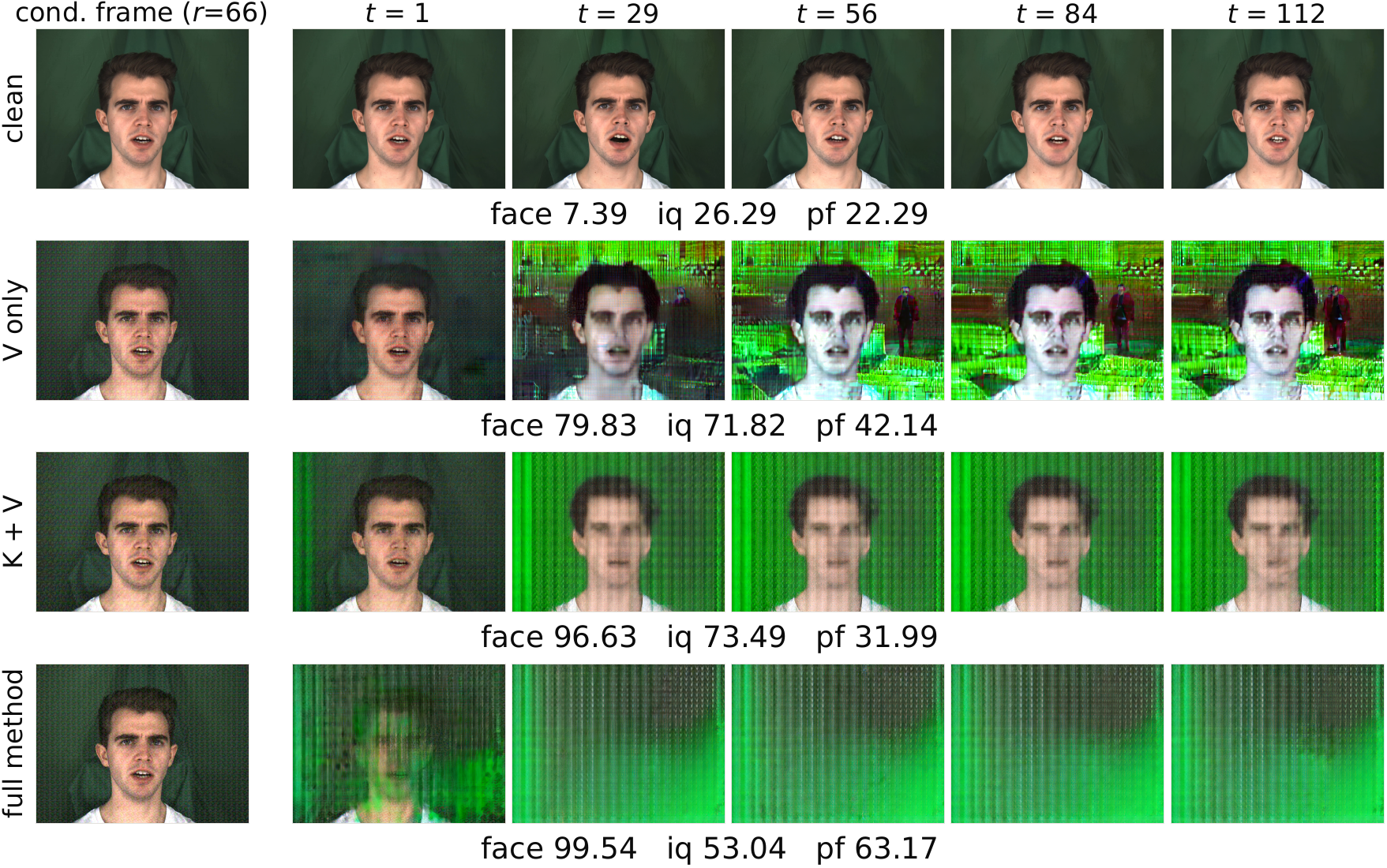}
\caption{Ablation of the Phase 1 protection objectives, adding K-attract, and auxiliary objectives to V-capacity collapse.}
\label{fig:abl1}
\end{figure}
For reference-based evaluation, we use a random reference frame
excluded from the background optimization set and evaluate on Wan2.2-TI2V-5B to show the effectiveness and HunyuanVideo-1.5 to show the transferability. For tuning-based evaluation, we finetune Wan2.2 with LoRA on the protected videos.

As shown in Table~\ref{tab:phase2_qualitative_eval}, the protected reference increases the Face Score on Wan2.2 from 16.79 to 82.66, indicating strong disruption of reference-based generation.
Our method also achieves a Face Score of 34.40 on the unseen HunyuanVideo1.5 model, and a Face Score of 51.09 on 1,500-step LoRA finetuning with Wan. These results confirm consistent identity disruption under both reference-based and LoRA-based customization.

\section{Additional Experiments}
\label{sec:exp}
\subsection{Ablation on Reference Capacity Collapse}
We progressively compare V-capacity collapse, its combination with K-attract,
and the full method. We use a held-out prompt excluded from EOT optimization and report Face Score, Image Quality (IQ) Score, and Prompt Following (PF) Score using the official Phase 1 evaluator released after the competition.

As shown in Figure~\ref{fig:abl1}, V-capacity collapse alone already causes substantial visual degradation, turning the person largely gray while still preserving a recognizable facial contour. After introducing K-attract, the remaining facial structure is further disrupted and the Face Score increases from 79.83 to 96.63.

These results suggest that V-capacity collapse and K-attract work jointly and form the core protection mechanism.
Adding auxiliary objectives further raises the Face Score to 99.54, providing complementary gains.

\subsection{Sensitivity to Target Identity}
\label{sensitive-target}

\begin{figure}[t]
    \centering
    \includegraphics[width=0.6\linewidth]{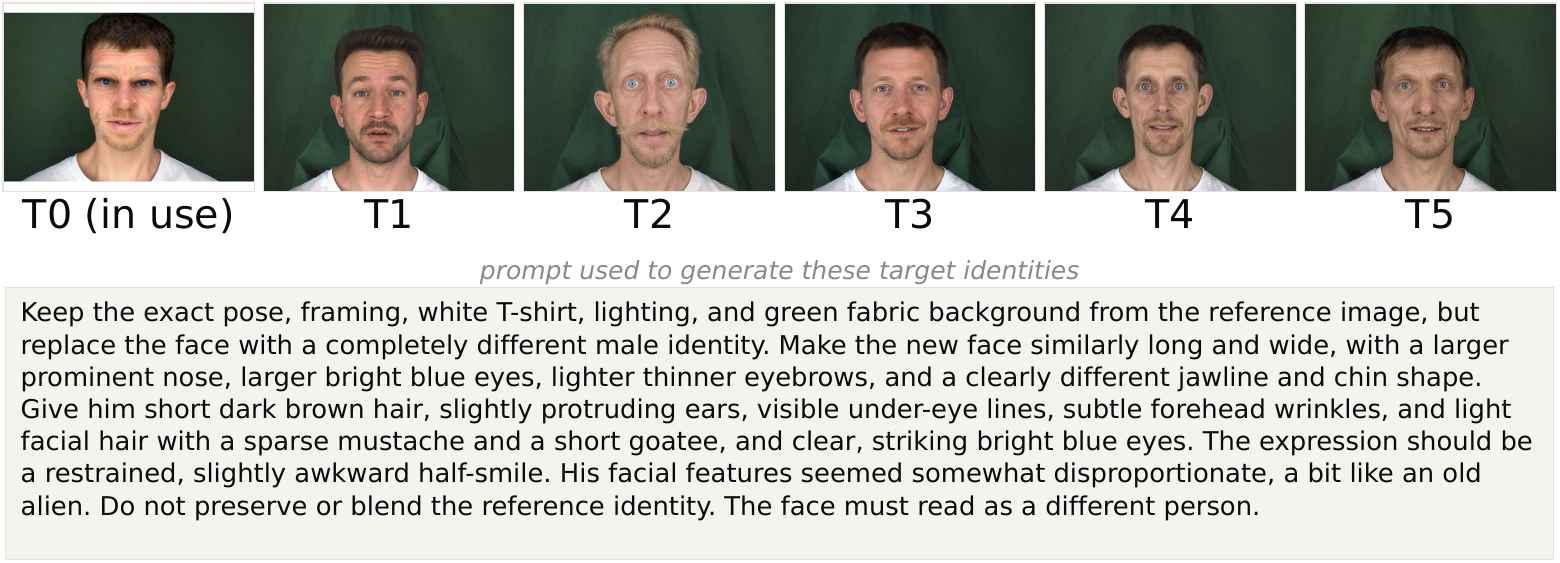}
    \caption{Candidate target identity images used for target selection}
    \label{fig:targets}
\end{figure}

We used ChatGPT Image\cite{openai_chatgpt} to generate multiple candidate images.
\begin{wraptable}[11]{r}{0.40\textwidth}
    \centering
    \setlength{\abovecaptionskip}{0pt}
    \caption{
        Comparison of different targets on HunyuanVideo-1.5
        and Wan2.2 LoRA finetuning 1500 steps.
        Results report the Face Score under the \textit{apple} prompt.
    }
    \label{tab:target_comparison}

    \resizebox{\linewidth}{!}{
        \begin{tabular}{lcc}
            \toprule
            \textbf{Target}
            & \textbf{Hunyuan}
            & \textbf{Wan2.2 LoRA} \\
            \midrule
            T0 (in use) & \textbf{66.86} & 93.12 \\
            T1          & 61.72 & 75.81 \\
            T2          & 60.56 & 87.27 \\
            T3          & 60.55 & \textbf{98.28} \\
            T4          & 62.49 & 83.60 \\
            T5          & 58.90 & 78.83 \\
            \bottomrule
        \end{tabular}
    }
\end{wraptable}
We investigate the sensitivity of the proposed face protection to the choice of target identity. Six independently generated target identities are evaluated under reference-based transferability on HunyuanVideo-1.5 and tuning-based effectiveness on Wan2.2-TI2V-5B with 1,500-step LoRA finetuning. The selected candidate target images are shown in Figure~\ref{fig:targets}, and their Face Scores are reported in Table~\ref{tab:target_comparison}. T0 is used in all main experiments and was selected using the development setting.

\subsection{Effectiveness against LoRA-based Customization}
\begin{figure*}[t]
    \centering

    \begin{minipage}[t]{0.43\textwidth}
        \centering
        \includegraphics[width=0.84\linewidth]{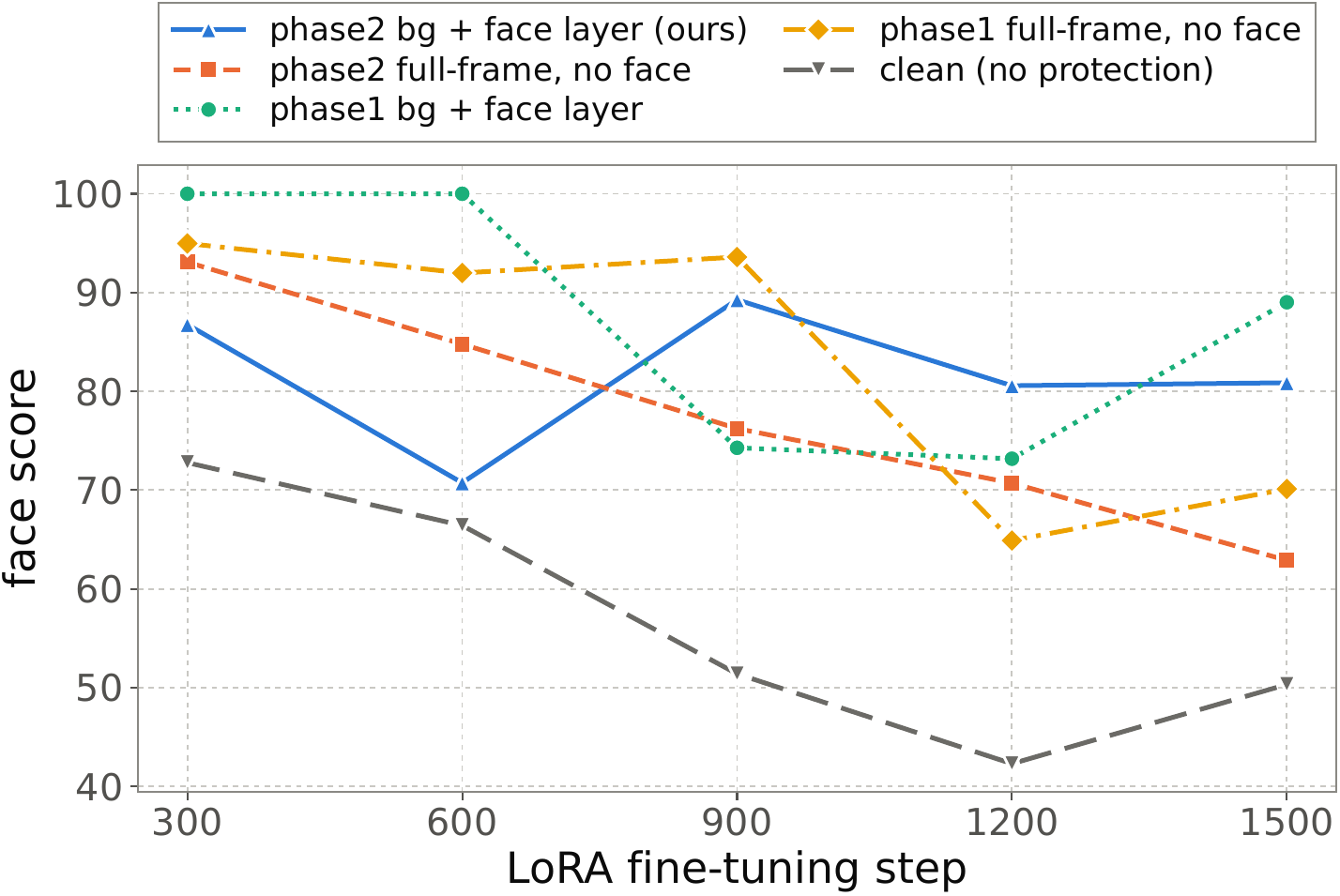}
        \captionof{figure}{Face Score across different Wan2.2 LoRA finetuning checkpoints for Phase 2 protection variants under the \textit{appear} prompt.}
        \label{fig:300step-facescore}
    \end{minipage}
    \hfill
    \begin{minipage}[t]{0.53\textwidth}
        \centering
        \includegraphics[width=0.84\linewidth]{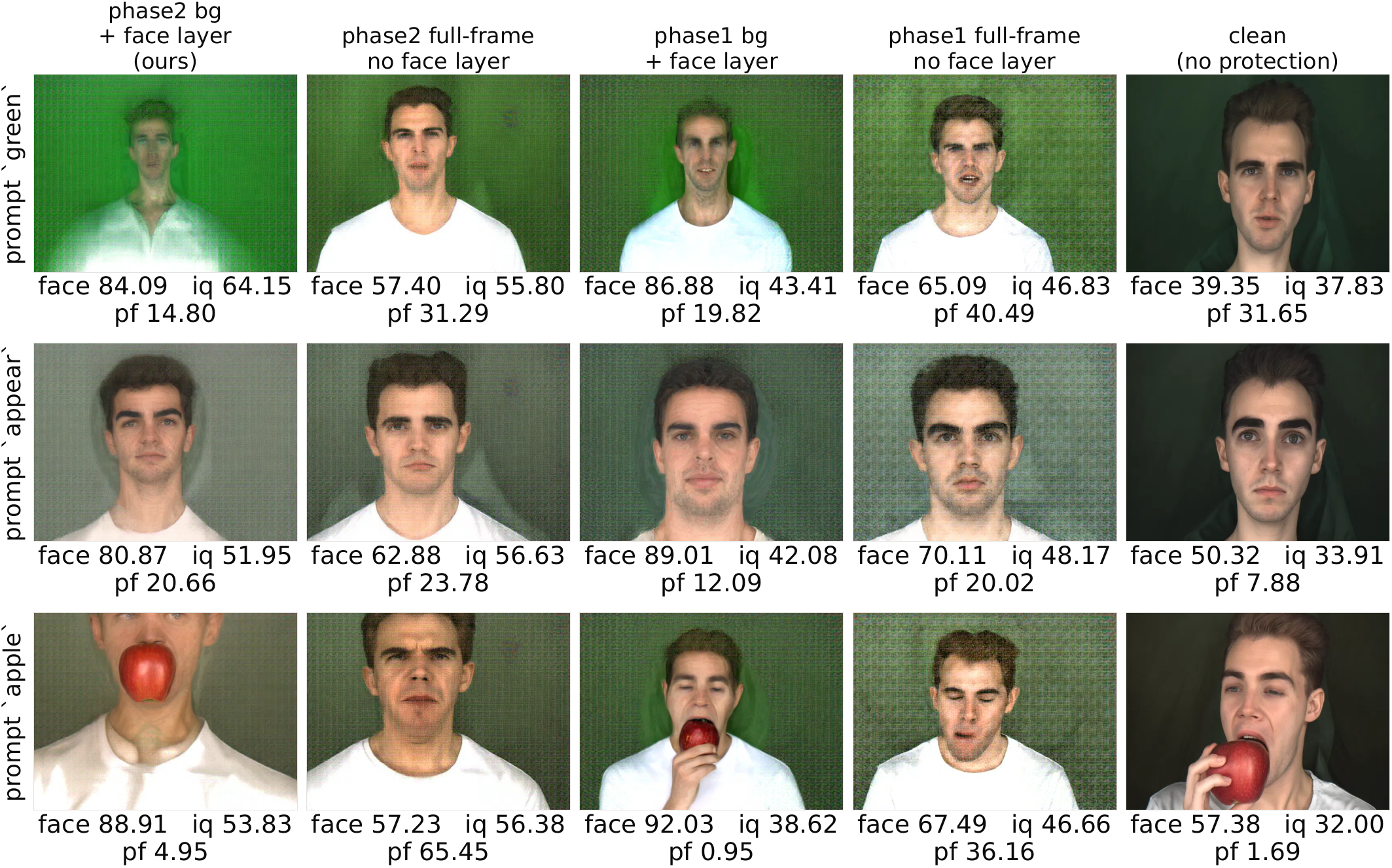}
        \captionof{figure}{Qualitative comparison of Phase 2 protection variants after 1,500-step Wan2.2 LoRA finetuning under three prompts: \textit{green, appear, apple}.}
        \label{fig:1500lora}
    \end{minipage}

\end{figure*}

We finetune Wan2.2-TI2V-5B for 1,500 steps using videos protected by different Phase 2 variants and save checkpoints every 300 steps. We report Face Scores using the \textit{appear} prompt in Figure~\ref{fig:300step-facescore}, and compare qualitative generations at 1,500 steps using the \textit{green}, \textit{appear}, and \textit{apple} prompts in Figure~\ref{fig:1500lora}. Detailed prompt settings are reported in Appendix~\ref{app:prompt}.

As shown in Figure~\ref{fig:300step-facescore}, variants with the Prior-Aligned Face Protection layer exhibit stronger persistence at later finetuning stages and clearly outperform their counterparts without it at 1,500 steps, suggesting persistent interference with identity learning. The background variant using the Phase 1 objectives achieves a slightly higher Face Score, while the proposed White-Box Background Protection causes stronger visual degradation, including subject blurring and out-of-frame faces. Overall, the face protection layer suppresses identity learning, while the Phase 2 background further disrupts generation quality.

\section{Conclusion}
\label{sec:conclusion}
Overall, we study identity protection in generative video customization and show that protective perturbations can effectively reduce an attacker's ability to reproduce the original identity under multiple threats, including reference-based and tuning-based customization. We hope this work provides insights for practical video identity protection and inspires stronger future defenses.

\section*{Acknowledgements}
We sincerely thank the competition organizers of the LifeGenIP Workshop @ ECCV 2026 for their well-prepared starting kit and timely responses to participants' questions, which greatly facilitated our work. We also thank Jian Yang, Bowen Zhang, and Mingchen Yin from Wuhan University for their participation and helpful discussions at the early stage of this competition.
\bibliographystyle{splncs04}
\bibliography{main}

\appendix
\section{Details of Auxiliary Objectives}
\label{aux}
Value-capacity collapse and Key attraction operate directly within the Transformer and serve as our core mechanism for weakening reference identity propagation. Building on these objectives, we further introduce a complementary Latent-level disruption, encouraging the protected reference to undergo a larger representation shift before entering the Transformer.

We observed that when V-capacity collapse and K-attract, further assisted by the auxiliary VAE disruption, become sufficiently strong, Wan2.2 can no longer reliably recover facial identity information from the protected reference. Under this weakened visual identity condition, the model increasingly relies on the remaining textual condition and its learned generative prior, often producing smooth and high-quality hallucinated faces unrelated to the original identity. This behavior already provides effective identity protection because the generated face no longer preserves the original identity. However, under the comprehensive Phase 1 evaluation, such model compensation may still retain favorable image quality and prompt following scores. Based on this observation, we introduce two auxiliary objectives to further suppress textual compensation and generative-prior compensation.

\subsection{Latent-level disruption.}
We enlarge the discrepancy between the protected reference and its clean counterpart in the VAE latent space. Let $z$ and $\tilde{z}$ denote the VAE latents of the clean and protected references, respectively. We define
\begin{equation}
\begin{aligned}
\mathcal{D}_{\mathrm{mse}}
&=
\frac{1}{|\mathcal{Z}|}
\left\|
\tilde{z}-z
\right\|_2^2,
\\
\mathcal{D}_{\mathrm{struct}}
&=
\mathcal{D}_{\mathrm{SSIM}}
\left(
\tilde{z},z
\right),
\\
\mathcal{L}_{\mathrm{VAE}}
&=
-\lambda_{\mathrm{mse}}\mathcal{D}_{\mathrm{mse}}
-\lambda_{\mathrm{struct}}\mathcal{D}_{\mathrm{struct}},
\end{aligned}
\end{equation}
where, $\mathcal{D}_{\mathrm{mse}}$ measures the overall representation shift between the protected and clean latents, while $\mathcal{D}_{\mathrm{struct}}$ measures their SSIM-based structural dissimilarity. Since both terms enter the minimization objective with negative signs, optimization jointly increases the latent-level MSE and structural discrepancy.

\subsection{Text-conditioned feature collapse.}
Similar to V-capacity collapse, we formulate this objective from the perspective of representation diversity. Under normal generation, different video tokens receive distinct text-conditioned representations according to their current states. We instead encourage their cross-attention outputs to become more similar, reducing the diversity of textual information delivered to different tokens. Let the cross-attention outputs at a target Transformer block be
\begin{equation}
O = \{o_i\}_{i=1}^{N}.
\end{equation}
We sample $J$ tokens from $O$ and maximize their mean pairwise cosine similarity:
\begin{equation}
\mathcal{L}_{\mathrm{text}}
=
-\frac{2}{J(J-1)}
\sum_{i<j}
\frac{
\langle o_i,o_j\rangle
}{
\|o_i\|_2 \|o_j\|_2
}.
\end{equation}
Minimizing this objective encourages the text-conditioned outputs of different video tokens to align in similar directions, thereby reducing the ability of cross-attention to provide differentiated textual conditions across generation positions.

\subsection{Image-quality degradation.}

Even when both visual and textual conditions are disrupted, pretrained video generation models still retain a strong facial generative prior. Without reliable reference identity, this prior often produces smooth and well-structured hallucinated faces that are unrelated to the original identity. Although such results do not recover the protected identity, they may weaken the protection effect against generation-quality degradation in Phase 1. To address this issue, we further introduce an image-quality objective based on natural scene statistics to suppress the high-quality compensation induced by the model prior.

We first compute the Mean-Subtracted Contrast-Normalized (MSCN) representation on the VAE reconstruction:
\begin{equation}
m(p)
=
\frac{
Y(p)-\mu(p)
}{
\sigma(p)+1
},
\end{equation}
where $Y$ denotes the luminance image, and $\mu(p)$ and $\sigma(p)$ denote the Gaussian local mean and local standard deviation, respectively. We then compute the normalized neighboring correlations of the MSCN coefficients along the horizontal, vertical, and two diagonal directions as the optimization objective:
\begin{equation}
\mathcal{L}_{\mathrm{IQ}}
=
-
\frac{1}{|\mathcal{D}|}
\sum_{\Delta\in\mathcal{D}}
\frac{
\mathbb{E}_{p}
\left[
m(p)m(p+\Delta)
\right]
}{
\mathbb{E}_{p}
\left[
m(p)^2
\right]+\xi
},
\qquad
\mathcal{D}
=
\{
\rightarrow,\downarrow,\searrow,\swarrow
\}.
\end{equation}

\section{Implementation Details}
\label{app:impl}

We detail the experimental settings of both phases below.

\subsection{Phase 1.}
We use Wan2.2-TI2V-5B as the white-box surrogate model and optimize a shared perturbation under an $\ell_\infty$ budget of $\epsilon=16/255$. We select $K=5$ representative frames using the motion-guided sampling strategy described in Sec.~\ref{sec:setting}. For EOT optimization, we draw $M=3$ samples per iteration and optimize for 90 iterations. The perturbation is initialized from $\mathcal{U}(-\epsilon/4,\epsilon/4)$, and we use a momentum-based sign update with step size $\eta=\epsilon/20$ and momentum coefficient $\beta=0.9$. The K-attract weight is set to $\lambda_K=0.5$. For the auxiliary objectives described in Sec.~\ref{aux}, the weights of the text-conditioned feature collapse, VAE latent MSE, VAE structural discrepancy, and image-quality objectives are set to $1.0$, $1.0$, $1.0$, and $2.0$, respectively. During optimization, we use 25 denoising steps and evaluate the attack objective at the fixed fifth denoising step. The Transformer blocks used for EOT are $\mathcal{B}=\{0,3,6\}$. Unless otherwise specified, all experiments use random seed 0.

\subsection{Phase 2.}
For Prior-Aligned Face Protection, we use the first frame of each clip as the anchor frame ($a=0$). The CIELAB channel weights are set to $\mathbf{w}=(1.2,1.0,1.0)$, the Gaussian smoothing standard deviation is $\sigma=6.0$ pixels, and the residual scaling coefficient is $\gamma=20.0$. The face masks are constructed from the union of facial regions detected across all frames, followed by dilation and feathering; the dilation size is $0.35$ times the face-box diagonal and the feathering standard deviation is $0.06$ times the mean face-box diagonal. For landmark-based alignment and temporal propagation, we use TPS regularization $\lambda_{\mathrm{TPS}}=10^{-3}$. For White-Box Background Protection, we retain the main Phase 1 settings, including $\lambda_K=0.5$, $M=3$, $\mathcal{B}=\{0,3,6\}$, $\eta=\epsilon/20$, and $\beta=0.9$, while increasing the optimization length to 200 iterations. We retain only V-capacity collapse, K-attract, and the two VAE latent disruption terms, with the two VAE weights both set to $1.0$. The Phase 2 EOT additionally includes H.264 recompression with CRF sampled from $[32,35]$, and JPEG quality is sampled from $[35,70]$.

\section{Prompt Settings}
\label{app:prompt}
We use three prompts throughout our experiments:
\begin{itemize}
    \item \textit{green}: ``the person (p3r5on) stands in front of a bright green screen and speaks to the camera''.
    \item \textit{appear}: ``the person (p3r5on) with short dark brown hair, thick eyebrows and light stubble looks straight at the camera''.
    \item \textit{apple} (used in Sec.~\ref{sensitive-target}): ``the person (p3r5on) raises a red apple to his mouth and takes a bite, chewing slowly''.
\end{itemize}
We use \textit{person} for reference-based customization and replace it with the rare token \textit{p3r5on} for tuning-based customization.

We use the \textit{green} prompt in the Phase 1 experiments, Phase 2 experiments, and the ablation on reference capacity collapse, while the sensitivity to target identity experiment uses the \textit{apple} prompt. For the qualitative evaluation of LoRA-based customization, we present results under all three prompts.

For the quantitative evaluation of Face Score across different Wan2.2-TI2V-5B LoRA fine-tuning checkpoints, we use the \textit{appear} prompt. We observe that, at the 600, 900, 1200 step checkpoints, the model trained on unprotected videos often fails to generate detectable faces under the \textit{green} and \textit{apple} prompts, suggesting that the target identity has not yet been reliably learned under these conditions. Face Score at these checkpoints therefore provides limited information about identity learning. In contrast, the \textit{appear} prompt can elicit a face from the model prior under textual guidance even at earlier checkpoints. As fine-tuning progresses, the Face Score on unprotected videos gradually decreases, indicating that the generated identity becomes increasingly aligned with the training subject. We therefore use the \textit{appear} prompt for the checkpoint comparison.

\end{document}